\documentclass[11pt]{article}
\usepackage{booktabs}
\usepackage[]{acl2023}
\usepackage{tabularx}
\usepackage{multirow}
\usepackage{multicol}
\usepackage{times}
\usepackage{latexsym}
\usepackage{graphicx}

\usepackage[T1]{fontenc}

\usepackage[utf8]{inputenc}

\usepackage{microtype}

\usepackage{inconsolata}

\usepackage[utf8]{inputenc}

\title{MahaSQuAD: Bridging Linguistic Divides in Marathi Question-Answering
}

\author{Ruturaj Ghatage$^{1,3}$, Aditya Kulkarni$^{1,3}$, Rajlaxmi Patil$^{1,3}$, Sharvi Endait$^{1,3}$, Raviraj Joshi$^{2,3}$ \\
        Pune Institute of Computer Technology, Pune$^1$ \\ Indian Institute of Technology Madras, Chennai$^2$\\L3Cube Labs, Pune$^3$}

\begin{document}
\maketitle

\begin{abstract}
Question-answering systems have revolutionized information retrieval, but linguistic and cultural boundaries limit their widespread accessibility. This research endeavors to bridge the gap of the absence of efficient QnA datasets in low-resource languages by translating the English Question Answering Dataset (SQuAD) using a robust data curation approach. We introduce MahaSQuAD, the first-ever full SQuAD dataset for the Indic language Marathi, consisting of 118,516 training, 11,873 validation, and 11,803 test samples. We also present a gold test set of manually verified 500 examples. Challenges in maintaining context and handling linguistic nuances are addressed, ensuring accurate translations. Moreover, as a QnA dataset cannot be simply converted into any low-resource language using translation, we need a robust method to map the answer translation to its span in the translated passage. Hence, to address this challenge, we also present a generic approach for translating SQuAD into any low-resource language. Thus, we offer a scalable approach to bridge linguistic and cultural gaps present in low-resource languages, in the realm of question-answering systems. The datasets and models are shared publicly at \url{https://github.com/l3cube-pune/MarathiNLP}.

\end{abstract}

\section{Introduction}

The Stanford Question Answering Dataset (SQuAD) \cite{rajpurkar2016squad} has emerged as a pivotal benchmark in the realm of natural language understanding and question answering and it has contributed to the development of cutting-edge machine learning models and techniques, enabling systems to provide accurate and contextually relevant answers to a wide array of questions in English. 
Question-answering systems have diverse applications in education, information retrieval, customer support, and more. These applications have the potential to bring significant benefits to Marathi speakers, enabling them to access knowledge and services in their native language. Question-answering systems are required for retrieving answers from a provided context, using the Extractive QA approach, thus leading to efficient retrieval of information from the questions provided.

However, as the world becomes increasingly interconnected and diversified, there is a growing need to make question-answering systems in natural language processing accessible and broadened across linguistic and cultural boundaries. Marathi is not just a language but a carrier of culture and identity for millions. It is important to recognize its cultural, regional, and linguistic significance. The creation of a Marathi question-answering dataset has the potential to impact a diverse population, both in India and among the Marathi-speaking community, enabling Marathi speakers to access and interact with information in their native language more effectively. Although the Indic language, Marathi, is the third highest spoken language in India, after Hindi and Bengali, it doesn’t have robust rich datasets in question answering despite these languages being spoken by several primarily in India, the country with the highest growing population \cite{joshi2022l3cube_mahanlp}.

While some efforts have been made to develop NLP resources for Marathi, there remains a substantial gap in terms of comprehensive question-answering datasets \cite{joshi2022l3cube_mahacorpus,pingle2023l3cube,patil2022l3cube,litake2022l3cube}. Existing research often lacks the depth and scale necessary for robust question answering. Our research builds upon and surpasses previous efforts by focusing on the SQuAD dataset, a gold standard in the field. Motivated by this challenge, our research aims to cover this linguistic divide by translating the SQuAD dataset into Marathi, using a robust approach. Our contributions can be summarised as follows:
\begin{enumerate}
    \item We provide a robust approach to retrieve translated answers within translated contexts with an adjustment in the span index accordingly.
    \item Creation of an exhaustive question-answering dataset, MahaSQuAD\footnote{\url{https://github.com/l3cube-pune/MarathiNLP}}, in the indic language Marathi, consisting of 118,516 individual data points in the training set,  11,873 data points in the validation set and 11,803 data points in the test set which has been translated from the SQuAD dataset, originally in English. 
    \item We curate a gold test set of 500 examples with manually verified answer spans. We release the MahaBERT\footnote{\url{https://huggingface.co/l3cube-pune/marathi-question-answering-squad-bert}} model finetuned on the L3Cube-MahaSQuAD dataset. 
\end{enumerate}
In the subsequent sections of this paper, we will provide a detailed account of our methodology for dataset translation, the challenges we encountered, the results, and the implications of our research for the field of natural language processing.
Our research extends beyond the immediate goal of creating a Marathi dataset. By addressing the linguistic diversity challenge, we hope to inspire similar efforts for other languages and demonstrate the potential for cross-linguistic research and development in the NLP community.

\section{Related Work}

In this section, we provide an overview of related work in question-answering datasets, machine translation for natural language processing, and multilingual natural language processing. Our discussion aims to contextualize our research and highlight the gaps in the existing literature that our proposed methodology addresses. \\
The Stanford Question Answering Dataset (SQuAD) has been a cornerstone in advancing the state of question-answering systems. \cite{rajpurkar2016squad} introduced the original English version of SQuAD, which has since served as a benchmark for evaluating machine comprehension models. Numerous studies have leveraged SQuAD for reading comprehension, question generation, and answer extraction \cite{devlin2018bert,liu2019roberta}. \\
To address the scarcity of datasets for new languages, methods have been explored to create datasets based on existing ones. Translating the context and answers within a dataset is a relatively straightforward task. However, a critical challenge arises when attempting to determine the exact index of the translated answer within the context, as the translated answer might not lie exactly in the same way as the answer in the context. This complexity stems from the translation of an answer, with or without surrounding context, which can vary, making it a hard task to directly pinpoint the index at which the answer resides within the context.\\
The majority of SQuAD-related research has focused on the English language, limiting its applicability in multilingual contexts. To our knowledge, few prior efforts have been made to translate SQuAD into multiple languages, and those that exist often lack detailed methodologies for ensuring high-quality translations \cite{wang2020reco} and also haven’t been in large numbers to ensure better training on the models. This gap underscores the importance of our work in extending SQuAD to a diverse set of Indic languages.\\
Machine translation has witnessed remarkable progress, with models like transformer-based architectures \cite{vaswani2017attention} significantly improving translation quality. However, translation in the context of natural language processing (NLP) poses unique challenges, including maintaining context, handling idiomatic expressions, and preserving semantic nuances.\\
Prior research has explored techniques to adapt machine translation for NLP tasks, such as paraphrasing \cite{mallinson2020zero}, context-aware translation \cite{platanios2019competence}, and domain-specific translation \cite{lample2019cross}. These studies have laid the foundation for addressing the translation challenges specific to question-answering datasets, which we further investigate in our proposed methodology.
Multilingual NLP has gained prominence due to its potential to foster cross-lingual communication and knowledge sharing. Researchers have explored methods for cross-lingual transfer learning \cite{artetxe2020cross}, zero-shot translation \cite{conneau2020unsupervised}, and multilingual model development \cite{devlin2018bert}. These efforts have broadened the scope of NLP applications across languages but have yet to always address the nuanced requirements of question-answering.\\
While SQuAD has been the benchmark for English question-answering, recent efforts have emerged to create multilingual question-answering datasets. For instance, the XQuAD dataset consisted of 240 paragraphs and 1190 question-answer pairs from SQuAD v1.1, which was translated by professional translators into ten different languages. \cite{artetxe2020cross} It expands the SQuAD concept to multiple languages, enabling the evaluation of question-answering models in a cross-lingual context. This work demonstrates the need and potential for multilingual question-answering resources, aligning with our objectives in the context of Indic languages. However, it consists of limited data points to evaluate, and it has focused on Indic languages such as Spanish, German, Greek, Russian, Turkish, Arabic, Vietnamese, Thai, Chinese, and Hindi, not consisting of Marathi, the language we aim to focus on.\\
\cite{kumar2022mucot} fine-tuned an mBERT-based QA model by translating and transliterating QA samples of the target language into other languages, increasing the performance of the models. \\
\cite{gupta2020bert} experiment with multilingual models on the task of Machine Comprehension, a part of Question-Answering (QA) for languages, Hindi and English. They experiment with mBERT on monolingual, zero-shot, and cross-lingual fine-tuning setups. \\
\cite{sabane2023breaking}, focuses on presenting the largest dataset Question answering dataset for Indic languages such as Hindi and Marathi, and they provide a performing model based on experimentation. We’ve aimed to provide a more robust and efficient technique for translating the dataset, that can provide an answer without the constraints of fixed word size in Marathi and with a larger dataset consisting of 130,319 data points.\\
Our research bridges the gap between machine translation and question answering in the multilingual context, proposing a novel methodology that combines the strengths of both fields to translate and adapt the SQuAD dataset effectively. 

\begin{figure*}[t]
    \centering
    \includegraphics[width= 12 cm]{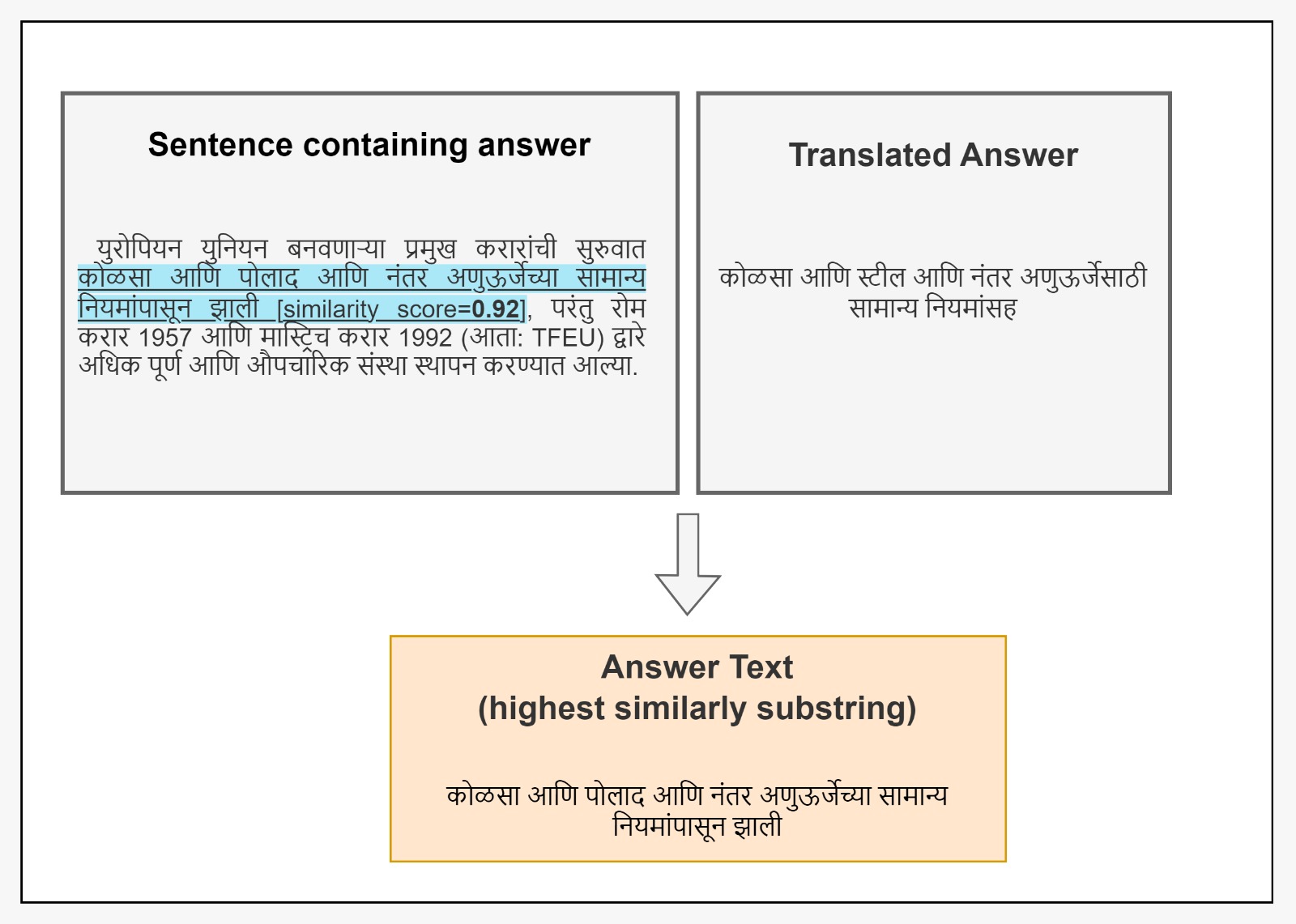}
    \caption{This figure illustrates the sentence containing the answer, the translated answer, and the answer text from the passage with has the highest similarity with the translated answer. }
    \label{fig:sentenceprocessing}
\end{figure*}

\section{Methodology}

In this paper, our approach encompasses a comprehensive translation of the entire English Question Answering dataset, SQuAD 2.0, into Marathi, resulting in the creation of MahaSQuAD. Employing a robust methodology, we not only achieve translation but also extract the translated answer and its corresponding answer span for each entry, complemented by transliteration. Furthermore, we meticulously construct distinct sets, including a training set, validation set, test set, and a gold set, each extensively examined in subsequent sections of the paper. To evaluate the efficacy of our translated dataset, we conduct training on both monolingual and multilingual models, encompassing MahaBERT, MahaROBERTA, mBERT and MuRIL-BERT. The detailed results of these training endeavors are systematically tabulated below, providing a comprehensive overview of our model performances.

\section{Experimental Setup}

\subsection{Data Collection}

We have used the Stanford Question Answering Dataset (SQuAD v2.0) dataset, originally in English, for translation into Marathi. Stanford Question Answering Dataset (SQuAD) is a reading comprehension dataset, consisting of Wikipedia articles, where the answer to every question is a segment of text, or span, from the corresponding reading passage, or the question might be unanswerable.

\subsection{Dataset Creation}

For dataset creation direct translation of a dataset item from English to any other language results in the following problems:
\begin{enumerate}
    \item The start index of an answer varies in the English context and in the translated context. In the English example, the answer starts at the 58th index, however, in Marathi the answer is found at the 50th index.
    \item The length of the answer may change after translation.
    \item When the answer text in the dataset is translated without surrounding context the translated answer may vary. In figure 1, \\
\texttt{\includegraphics[height=2 cm]{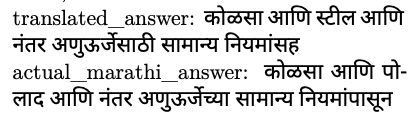}} \\
There is an actual difference between the answer present in the translated context and in the answer translated directly.

\end{enumerate}

To translate the dataset accurately and resolve the above issues we propose a robust solution:
\begin{enumerate}
    \item Divide the context into sentences.
    \item Identify the English sentence containing the answer.
    \item Translate the English sentence and answer into Marathi.
     \item Find the phrase in the Marathi context sentence that has the highest similarity with the translated answer. 
    \item  Set the phrase as the new answer.
\end{enumerate}


Firstly, we translate the question and the title from the entity using DeepTranslator. In our approach, we divide the English context into sentences using the nltk library and extract the sentence that contains the answer. Each sentence is translated individually. These sentences are used to create the translated context. Then sentence containing the answer and the answer is translated. The translated answer is then compared with all possible substrings present in the translated sentence. We iterate over all phrases of varying lengths in sentences calculate the similarity of each substring with the answer and store it in a 2D matrix. The similarity score is calculated using the SimilarityAnalyzer from the MahaNLP library. From this matrix, we find the substring with the maximum similarity score and set it as the base answer. Then we append adjacent words to the base answer and check the similarity score of the new phrase while allowing a threshold value for the score to be 1\% less than maximum similarity. This makes the solution more dynamic and ensures that no part of the original answer is missing in the translated answer.  
Lastly, we transliterate over the entire processed Marathi dataset to convert named entities into the Devanagari Script to ensure uniformity in the script and the data, using the AI4Bharat Transliteration engine.
\begin{figure}[t]
\centering
    \includegraphics[width= 5 cm]{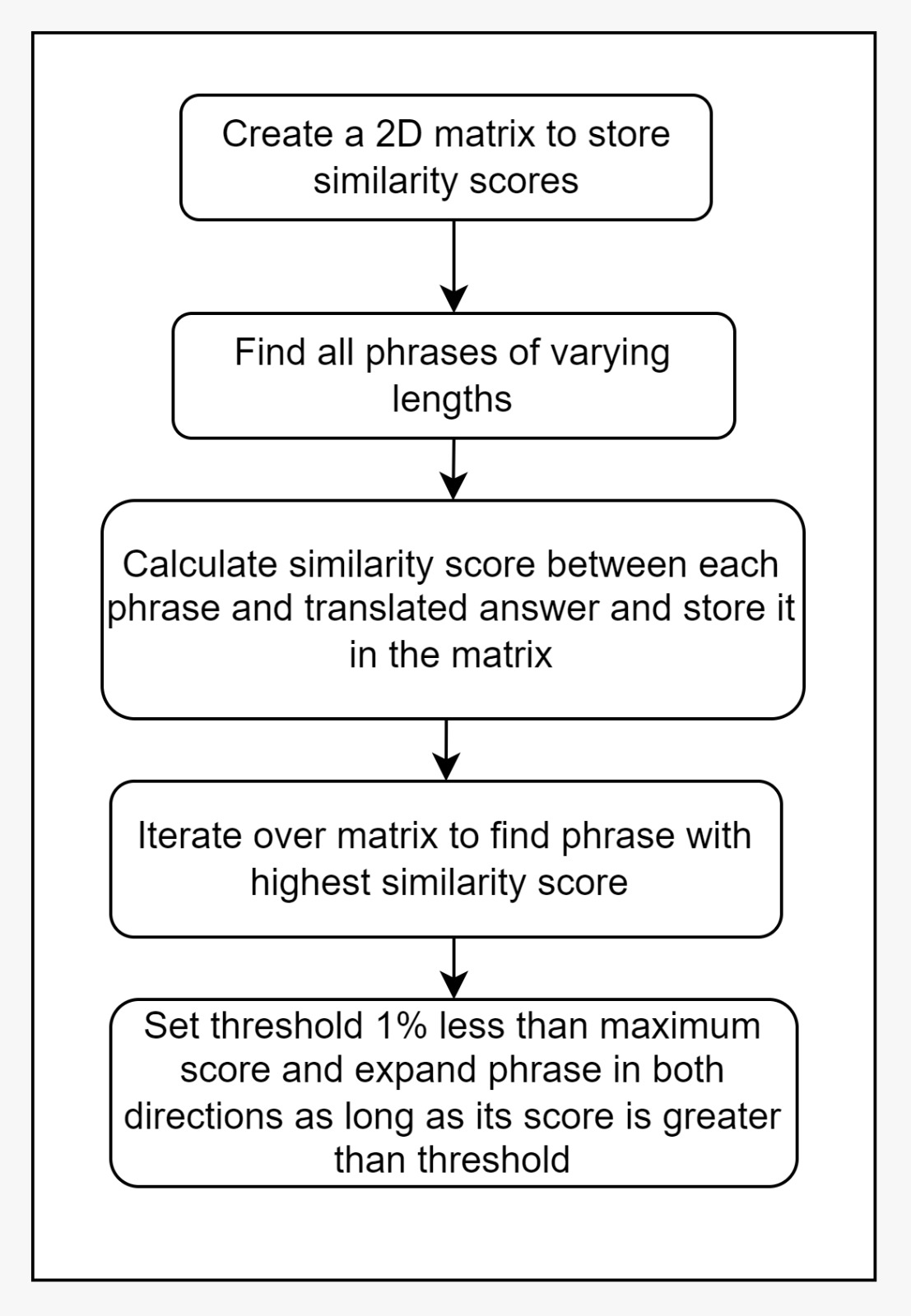}
    \caption{Algorithm for obtaining the answer and the answer span from the context}
    \label{fig:sentenceprocessing}
\end{figure}
\begin{figure*}[t]
    \centering
    \includegraphics[width= 16 cm]{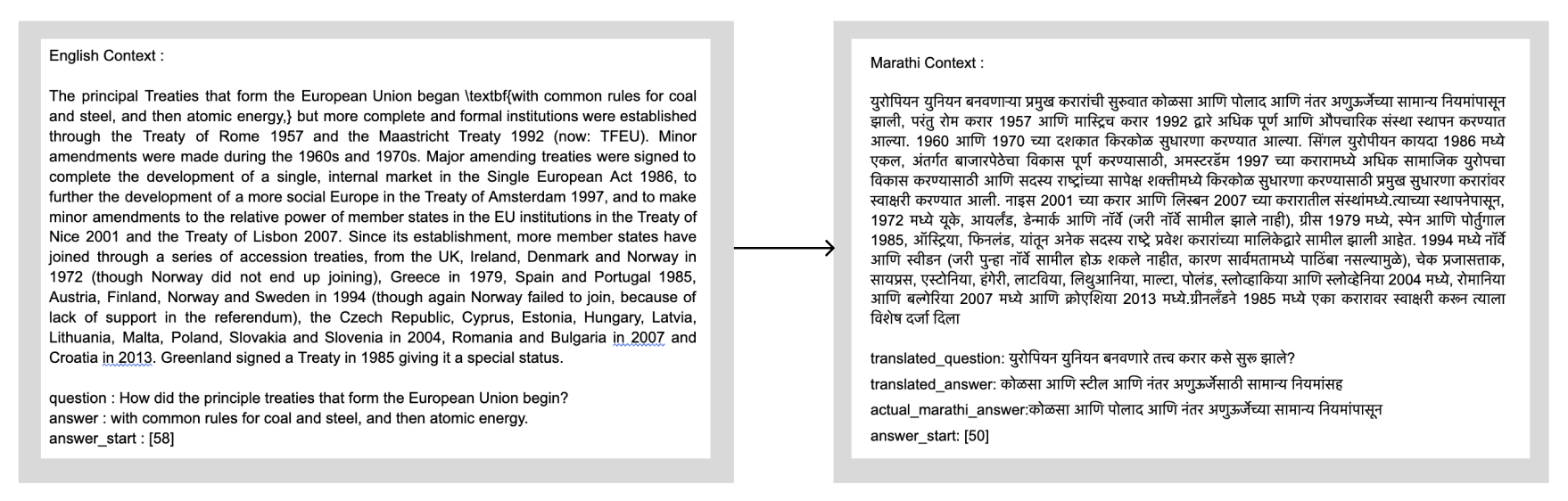}
    \caption{Left: An English SQuAD example, Right: Corresponding entry from MahaSQuAD}
    \label{fig:sentenceprocessing}
\end{figure*}

\subsection{Challenges during Preprocessing}
During the process of curating our dataset, there were certain problems which we faced. Some of these problems required innovative approaches, while some of them custom These are a list of the problems and the solutions we made use of are as follows:      
\begin{enumerate}
    \item Camel Case words: \\
During our translation process, we encountered numerous instances of CamelCase words. CamelCase is a naming convention that capitalizes each word within a compound term without the use of hyphens or spaces. Some prominent examples that we ran into include 'YouTube,' 'iPad,' and 'McDonald's.' These CamelCase words presented a significant challenge during the translation process, impacting both accuracy and readability. To address this, we simply converted the entire CamelCase word to lowercase. This facilitated a smoother and more precise translation process. 
\item Abbreviations: \\
We came across multiple instances of abbreviations in our dataset. These abbreviations posed a difficult problem. Our novel approach in the translation process included splitting the sentences using full stops. But many times, abbreviations like “L.A”, and ”L . A .” were getting split into two sentences like 'L' and ‘A’. This had an adverse effect on our translation. Our solution involved using the NLTK sentence tokenizer to distinguish between abbreviations and sentence boundaries. This streamlined the translation process and ultimately enhanced the extraction of precise answers.
\item English Numbers to Devanagari Numbers: \\
The predominant challenge we encountered was the failure of English numbers to be translated into Devanagari numerals. In numerous examples, numerical values remained unaltered during the translation process. To address this issue, we implemented the AI4Bharat Transliteration Engine, ensuring that all the numbers were transliterated into their Devanagari counterpart.
\item Special English Characters: \\
We encountered numerous examples with special English characters, such as 'é'.  Given the absence of corresponding Marathi characters, we employed the AI4Bharat Transliteration Engine to facilitate the transliteration process. This approach ensured that the majority of the examples contained minimal to no English characters after transliteration. This enhanced the target language's consistency and helped the translation process dearly.
\end{enumerate}

\subsection{Dataset Statistics}

MahaSQuAD contains all the samples from SQuAD 2.0 translated in Marathi using our robust approach using a machine translation model for Indic Languages. We use 118,516 samples for the training set, 11873 samples for the validation set, 11,803 samples for the test set, and 500 examples for the Gold set. Each data point is divided into title, context, question, answer, and answer start. The availability of the MahaSQuAD dataset makes the way for research projects in linguistics, and language technology thus, offering new opportunities for academics and institutions to develop and enhance Marathi language resources.

\begin{table}[h]
    \centering
    \begin{tabular}{|c|c|}
    \hline
         Training set & 118,516 samples \\
         Validation set & 11,873 samples \\
         Test set & 11,803 samples\\
         Gold set & 500 samples \\
         \hline
    \end{tabular}
    \caption{Dataset Statistics}
    \label{tab:my_label}
\end{table}

\section{Experimentation}
  \begin{table*} [t]
    \centering
    \small\begin{tabular}{|ccccccccc|}
    \hline
    
   Model & EM\% & F1\% & EM & F1 & EM & F1 & BLEU\% & BLEU\% \\ [0.5 ex]
    & & &(Has\_ans) & (Has\_ans)& (No\_ans) & (No\_ans)&(Unigram)& (Bigram) \\
    \hline 
  
    MahaBERT & 51.28 & 54.88 & 51.04 & 58.31 & 51.52 & 51.52 & 57.9 & 49.9\\
    MahaRoBERTa  & 55.33 & 58.79 & 49.54 & 56.54& 60.98 & 60.98 & 57.5 & 48.1\\
     mBERT & 60.08 & 61.53 & 40.76 & 43.71& 78.93 & 78.93 & 58.3 & 50.5\\
    MuRIL-BERT & 50.13 & 53.91 & 51.26 & 58.92& 49.03 & 49.03 & 57.7 & 49.4\\
   
    \hline
    \end{tabular}
    \caption{Exact Match (EM), F1, BLEU (Unigram), BLEU (Bigrams) for various models on MahaSQuAD}
    \label {table:1}
    \end{table*}

       \begin{table*} [t]
    \centering
    \small\begin{tabular}{|ccccccccc|}
    \hline
    
   Model & EM\% & F1\% & EM & F1 & EM & F1 & BLEU\% & BLEU\% \\ [0.5 ex]
    & & &(Has\_ans) & (Has\_ans)& (No\_ans) & (No\_ans)&(Unigram)& (Bigram) \\
    \hline 
  
    MahaBERT & 55.8 & 61.86 & 33.33 & 45.65 & 77.56 & 77.56 & 41.6 & 32.7\\
    MahaRoBERTa  & 59.6 & 64.91 & 18.69 & 29.5 & 99.21 & 99.2 & 61.1 & 54.6\\
     mBERT & 52.4 & 54.85 & 10.16 & 15.14 & 93.3 & 93.3 & 47.0 & 39.9\\
    MuRIL-BERT & 56 & 61 & 28.86 & 39.16 & 82.28 & 82.28 & 42 & 33.9\\
   
    \hline
    \end{tabular}
    \centering
          \caption{Exact Match (EM), F1, BLEU (Unigram), BLEU (Bigrams) for various models on MahaSQuAD for the Gold Testset}
    \label {table:2}
    \end{table*}
\subsection{Model Selection}
Following are the Language Models used for training the dataset:
\begin{enumerate}
    \item \textbf{MahaBERT} \\
This model was developed in \cite{joshi2022l3cube_mahacorpus,joshi2022l3cube_hindbert}. This model is derived from fine-tuning the multilingual BERT case model. As the name suggests, it is mainly trained in the Marathi language. This model was mainly trained on the L3-Cube MahaCorpus Dataset which is a Marathi Monolingual Dataset along with various other open-sourced Marathi monolingual datasets.
\item \textbf{MahaRoBERTa} \\
It is a multilingual RoBERTa (xlm-roberta-base) model fine-tuned on L3Cube-MahaCorpus and other publicly available Marathi monolingual datasets.
\item \textbf{mBERT} \\
mBERT, or Multilingual Bidirectional Encoder Representations from Transformers, is a powerful language model developed by Google that can understand and interpret text across 104 languages. By using the transformer architecture it can analyze and understand the meaning of the text by considering the context in which it appears.
\item \textbf{MuRIL-BERT}\\
It stands for "Multilingual Representations for Indian Languages," and "MURIL-BERT" is a variant of the BERT (Bidirectional Encoder Representations from Transformers) model specifically designed for understanding and processing text in various Indian languages. It was developed to handle the linguistic diversity and complexity of the Indian subcontinent, providing advanced natural language understanding capabilities for languages like Hindi, Bengali, Tamil, and more. MURIL-BERT is a valuable tool for NLP (Natural Language Processing) tasks in Indian languages, including sentiment analysis, language translation, and content generation.
\end{enumerate}

\subsection{Gold Testset}
For our research objectives, we conducted a random selection of 500 samples from the test dataset. The Gold test set comprises answers and answer spans determined through manual calculation. This was done to give us a comprehensive idea of how accurate our approach is when compared with manual calculations for the same samples which were been evaluated using our robust approach. We manually calculated the answer and the answer span for each example through the following steps:
\begin{enumerate}
    \item For each example, we pinpointed the sentence containing the answer.
    \item Corresponding to each question, we obtained the correct answer from the sentence that was pinpointed in the previous step.
    \item From the extracted answer, we precisely located the answer span for each example. Specifically, we determined the starting index of the answer within the sentence.
\end{enumerate}

\subsection{Experimental Settings}
We conducted fine-tuning on our models using a custom dataset, spanning three epochs and utilizing A100 GPUs with a consistent batch size of 8. The carefully selected hyperparameters include n\_best\_size ( which refers to the number of predictions provided per question ) set to 2, significantly shaping the training dynamics and influencing the experimental outcomes. The other key hyperparameters employed during fine-tuning included a learning rate of 4e-5 and the AdamW optimizer. These adjustments played a crucial role in refining the model and enhancing its performance.








\section{Results}

In this study, we have used various monolingual and multilingual models in
Marathi. We have used the following evaluation metrics: 

 \begin{enumerate}
     \item \textbf{EM (Exact Match)}:

The Exact Match (EM) score in NLP measures the accuracy of question-answering systems by assessing whether their generated answers are identical, character by character, to the reference answers. It provides a binary metric, where a 1 indicates an exact match (correct answer) and a 0 indicates a mismatch (incorrect answer). The EM score is a useful measure for evaluating the precision of QA systems.

\item \textbf{F1 (F-beta score)}:

The F1 score in NLP is a single metric that balances a model's precision (accuracy of positive predictions) and recall (coverage of actual positives). It is commonly used to evaluate the overall performance of classification and information retrieval systems, providing a harmonized measure of a model's effectiveness. The F1 score ranges from 0 to 1, with higher values indicating better performance.

\item \textbf{BLEU Score (Bilingual Evaluation Understudy)}:
BLEU (Bilingual Evaluation Understudy) is a metric for automatically evaluating machine-translated text. The BLEU score is a number between zero and one that measures the similarity of the machine-translated text to a set of high-quality reference translations.
 \end{enumerate}

From the results of Table 2, the monolingual models MahaBERT and MahaROBERTa outperform the multilingual models mBERT and MuRIL-BERT in terms of all the given parameters. MahaBERT and MahaROBERTa excel in exact match accuracy because they are fine-tuned and they deeply understand the nuances of that language. They benefit from language-specific training and preprocessing, resulting in precise answers and exact matches. While mBERT demonstrates higher EM and F1 scores compared to other models, it is not inherently superior. This is because it often predicts answers to be empty, which boosts its overall EM and F1 score but results in lower scores when considering samples containing answers.

A drawback of this dataset is its reliance on an automated data creation process, in contrast to the manually curated gold test set. Even for the gold dataset, both the monolingual models (MahaBERT and MahaRoBERTa) outperform the performance of multilingual models referring to Table 3. Consequently, there is a possibility of minimal noise being introduced.

\section{Future Scope}

Our proposed pipeline has showcased enhanced efficiency in the creation of translated SQuAD datasets. This improved pipeline can be extended to cater to other low-resource languages, facilitating the generation of question-answer datasets for those languages as well. Additionally, leveraging the accuracy in identifying the most relevant spans can be applied to generate datasets for various other applications.

\section{Conclusion}

In this research, our objective was to address the scarcity of efficient question-answering (QnA) datasets in low-resource languages, specifically focusing on Marathi. Our central research question, "Can the English Question Answering Dataset be effectively translated into Marathi to foster linguistic accessibility?" guided our investigation into this crucial domain. Employing a rigorous and robust methodology, we successfully translated the entire English Question Answering Dataset into Marathi, resulting in the creation of the 'MahaSQuAD' dataset.

Our findings highlight the potential of the 'MahaSQuAD' dataset to empower Marathi speakers, providing accessibility to technology and knowledge in their native language. Furthermore, by introducing a scalable approach for translating QnA datasets into low-resource languages, we pave the way for enhanced information retrieval in Marathi.

Notably, this dataset serves as a valuable benchmark dataset for developing Marathi question-answering models. Its applicability extends beyond research, proving useful in creating applications such as chatbots for low-resource languages. The notable success of MahaBERT and MahaROBERTa, as evidenced by state-of-the-art results, emphasizes the efficacy of these models in the realm of question-answering datasets.

\section*{Acknowledgment}
This work was done under the L3Cube Pune mentorship
program. We want to thank our mentors at L3Cube for their continuous support and
encouragement. This work is a part of the L3Cube-MahaNLP
project \cite{joshi2022l3cube_mahanlp}.

\bibliography{main}
\bibliographystyle{acl_natbib}

 \end{document}